\documentclass[letterpaper, 10 pt, conference]{ieeeconf}  

\IEEEoverridecommandlockouts                              

\usepackage{graphicx}
\usepackage{amsmath,amssymb,amsfonts}
\usepackage{algorithmic}
\usepackage{algorithm}
\usepackage{xcolor}
\usepackage{animate}
\usepackage{subfig}
\usepackage{tabularx,booktabs}
\usepackage{url}

\title{\LARGE \bf
Fusing Event-based Camera and Radar for SLAM Using Spiking Neural Networks with Continual STDP Learning
}

\author{Ali Safa$^{1,4}$, Tim Verbelen$^{2,4}$, Ilja Ocket$^{4}$, André Bourdoux$^{4}$, Hichem Sahli$^{3,4}$, \\
Francky Catthoor$^{1,4}$, Georges Gielen$^{1,4}$
\thanks{This research has received funding from the Flemish Government (AI Research Program) and the European Union's ECSEL Joint Undertaking under grant agreement n° 826655 - project TEMPO.}
\thanks{$^{1}$ Faculty of Electrical Engineering (ESAT) KU Leuven, 3001, Belgium}%
\thanks{$^{2}$ IDLab, Gent University, B-9052 Gent, Belgium}%
\thanks{$^{3}$ ETRO, VUB, 1050 Brussels, Belgium}
\thanks{$^{4}$ imec, Kapeldreef 75, 3001, Leuven, Belgium
        {\tt\small \{Ali.Safa, Tim.Verbelen, Ilja.Ocket, Andre.Bourdoux, Hichem.Sahli, Francky.Catthoor\}@imec.be},
        {\tt\small Georges.Gielen@kuleuven.be}}%
}

\begin{document}

\maketitle
\thispagestyle{empty}
\pagestyle{empty}

\begin{abstract}
This work proposes a first-of-its-kind SLAM architecture fusing an event-based camera and a Frequency Modulated Continuous Wave (FMCW) radar for drone navigation. Each sensor is processed by a bio-inspired Spiking Neural Network (SNN) with continual Spike-Timing-Dependent Plasticity (STDP) learning, as observed in the brain. In contrast to most learning-based SLAM systems
, our method does not require any offline training phase, but rather the SNN continuously learns features from the input data on the fly via STDP. At the same time, the SNN outputs are used as feature descriptors for loop closure detection and map correction. We conduct numerous experiments to benchmark our system against state-of-the-art RGB methods and we demonstrate the robustness of our DVS-Radar SLAM approach under strong lighting variations.

\end{abstract}

\section*{Multimedia material}
Please watch a demo video of our SNN-based DVS-Radar fusion SLAM at \url{https://youtu.be/a7gvZWNHGoI}

\section{Introduction}
Simultaneous Localisation and Mapping (SLAM) is an important problem for autonomous agents such as drones \cite{ultimateslam, latentslam}. Most state-of-the-art SLAM systems integrate raw odometry data with feature matching using standard RGB cameras in order to detect \textit{loop closures} (i.e., places already visited by the agent) and to correct the drift in raw odometry accordingly \cite{orbslam2}. However, using RGB cameras alone does not provide upmost robustness, such as insensitivity to lighting variations and environmental conditions \cite{ultimateslam}. Therefore, multi-sensor SLAM systems fusing RGB with e.g., lidar, radar and event-based cameras have emerged with increased robustness compared to RGB alone \cite{latentslam, sensorfusionslam}.
\setcounter{figure}{1}
\begin{figure}[htbp]
\centering
    \includegraphics[scale = 0.49]{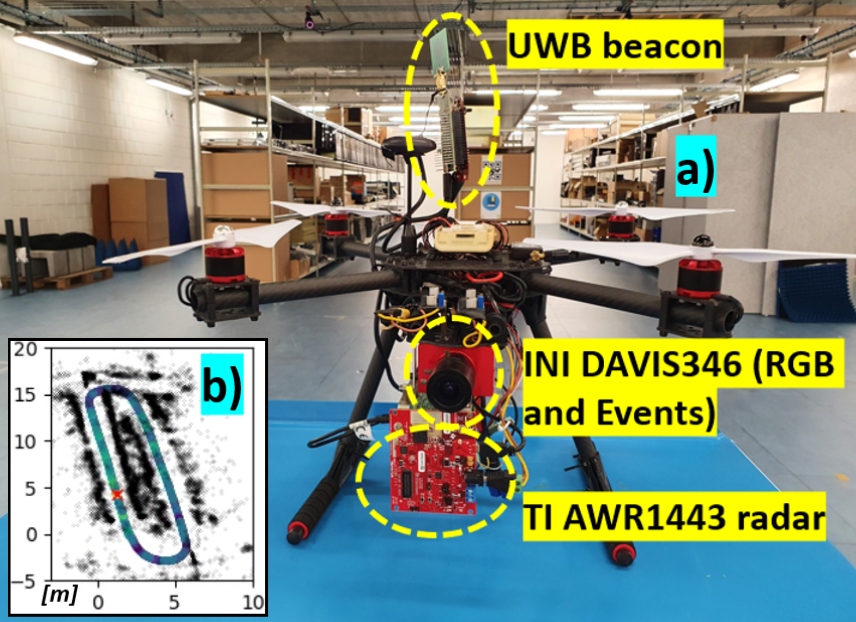}
    \caption{\textit{\textbf{a) Our sensor fusion drone setup} used for jointly acquiring event camera and radar data. The warehouse environment (detailed in \cite{indooruwb}) in which the drone navigates is also shown. The warehouse is equipped with Ultra Wide Band (UWB) localisation for ground truth acquisition \cite{indooruwb}. \textbf{b)} Event and radar data are fed to SNNs that continually learn on the fly via STDP (see Fig. \ref{fusionsnn}). The SNN output is used for loop closure detection to perform SLAM. Radar detections are also used to model the obstacles (black dots in \textbf{b}).}}
    \label{droneplacehold}
\end{figure}

Event-based cameras (also called \textit{Dynamic Vision Sensors} or DVS) are a novel type of imaging sensor composed of independent pixels $x_{ij}$ that asynchronously emit spikes whenever the change in light log-intensity $|\Delta L_{ij}|$ sensed by the pixel crosses a certain threshold $C$ \cite{gallego} (see Fig. \ref{dvs_concept}). In contrast to RGB cameras, DVS cameras can still perform well in low-light conditions \cite{ultimateslam} and produce a \textit{spatio-temporal} stream that contains patterns in both the spatial and spike timing dimensions, making them a natural choice as input for \textit{Spiking Neural Networks} (SNNs). Indeed, SNNs are bio-plausible neural networks that make use of spiking neurons, communicating via binary activations in an event-driven manner, matching the DVS principle \cite{snnintro}. 
\setcounter{figure}{0}
\begin{figure}[b!]
\centering
    \includegraphics[scale = 0.43]{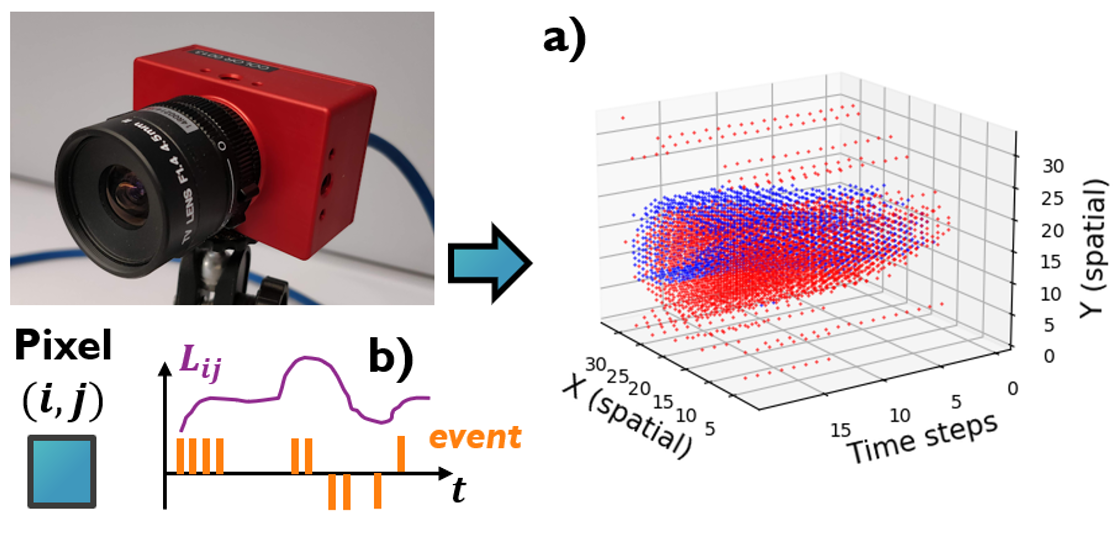}
    \caption{\textit{\textbf{Conceptual illustration of event-based vision.} a) The camera outputs a stream of spikes in both space and time. 
    b) Each pixel $x_{ij}$ fires a spike when its change in light log-intensity $\Delta L_{ij}$ crosses a threshold. The spike is positive when $\Delta L_{ij} > 0$ (red dots) and negative (blue dots) otherwise.}}
    \label{dvs_concept}
\end{figure}
\setcounter{figure}{2}

SNNs have recently gained huge attention for ultra-low-energy and -area processing in power-constrained edge devices such as drones \cite{yulia}. In addition, the use of event-driven learning rules such as Spike-Timing-Dependent Plasticity (STDP) \cite{Bi10464} enables \textit{unsupervised} learning at the edge using emerging \textit{sub-milliwatt} neuromorphic chips \cite{stdpchip,loihi}, as opposed to power-hungry GPUs. Furthermore, SNN-STDP systems are also a good choice for continual, \textit{on-line} learning \cite{yuliaeditor}, dropping the need for offline SNN training. Hence, this work studies the use of SNNs equipped with STDP learning for fusing DVS and radar data on power-constrained drones.

The use of radar sensing for drone navigation has recently been investigated in a growing number of works \cite{radarmotion, radarrgbdataset, radarobstav, mydetector, ali}. Indeed, fusing radar sensing with vision-based sensors such as DVS is attractive since radars intrinsically provide complementary information to camera vision, such as target velocity and position in a range-azimuth map (also called \textit{bird's eye view} vs. projective plane in cameras) \cite{ali}. In addition, radars are robust to environmental conditions as they are not sensitive to occlusion by dirt and can sense in the dark \cite{radarrgbdataset}.

This work deviates from most fusion-based SLAMs by proposing a \textit{first-of-its-kind}, bio-inspired SLAM system fusing event camera with radar (see Fig. \ref{droneplacehold}, \ref{fusionsnn}), using SNNs that \textit{continuously} learn via STDP, as observed in the brain \cite{Bi10464}. 
It also deviates from most learning-based SLAM systems which typically require the \textit{offline} training of a Deep Neural Network (DNN) on a dataset of the working environment captured beforehand \cite{latentslam}. In contrast, our DVS-Radar fusion SNN learns on the fly and keeps adapting its weights via \textit{unsupervised} STDP as the drone explores the environment. At the same time, the SNN outputs are fed to a bio-inspired RatSLAM back-end \cite{ratslam} for loop closure detection and map correction. 
Crucially, our continual STDP learning approach enables the deployment of our system in environments \textit{not} captured by datasets and therefore, not known \textit{a priori} (vs. offline training in state-of-the-art DNN-based SLAMs \cite{latentslam}).


We use our sensor fusion drone shown in Fig. \ref{droneplacehold} 
to jointly acquire DVS and radar data during multiple drone flights in a challenging, indoor environment, in order to perform SLAM (see Fig. \ref{droneplacehold} b). We assess the performance of our proposed DVS-Radar SNN-STDP SLAM system against ground truth positioning, recorded via \textit{Ultra Wide Band} (UWB) beacons \cite{indooruwb}. The main contributions of this paper are the following:
\begin{enumerate}
    \item We propose what is, to the best of our knowledge, the first continual-learning SLAM system which fuses an event-based camera and an FMCW radar using SNNs.
    \item We propose a method for radar-gyroscope odometry, where radar sensing provides the drone's velocity, and a method for obstacle modelling via radar detections.
    \item We experimentally assess the performance of our SLAM system on three different flight sequences in a challenging warehouse environment and we show the robustness of our system to strong lighting variations.
\end{enumerate}

This paper is organized as follows. Related works are discussed in Section \ref{related}, followed by background theory covered in Section \ref{preli}. Our proposed methods are presented in Section \ref{props}. Experimental results are shown in Section \ref{expres}. Conclusions are provided in Section \ref{concs}.

\section{Related works}
\label{related}
A growing number of bio-inspired \cite{ratslam,slamsnn1,slamsnn2} and sensor fusion \cite{ultimateslam, latentslam, sensorfusionslam} SLAM systems have been proposed in recent years . Among the most related to this work, a DVS-RGB SLAM system has been proposed in \cite{ultimateslam}, providing robust state estimations by fusing event-based cameras, RGB and raw IMU odometry. In addition, the system of \cite{ultimateslam} has been implemented on a drone for indoor navigation and was shown to be robust to drastic changes in lighting conditions and in low-light scenarios. In contrast to \cite{ultimateslam}, which makes use of hand-crafted features, our system uses an SNN with STDP learning. In addition, we do not fuse the richer information obtained from RGB as in \cite{ultimateslam}, but rather fuse DVS with \textit{sparser} radar detections instead, which makes the reliable template matching of the SNN outputs challenging. 

Recently, the \textit{LatentSLAM} system has been proposed in \cite{latentslam} as a learning-based pipeline using a DNN encoder which provides \textit{latent codes} for template matching, trained \textit{offline} on a dataset capturing the environment in which the robot must navigate. The inferred latent codes are fed to the loop closure detection and map correction back-end of the popular \textit{RatSLAM} system \cite{ratslam} to correct the drift in raw odometry. 
In contrast, our proposed continual learning system \textit{does not} require any offline training phase, enabling its deployment in \textit{unseen environments} without the requirement of capturing a dataset of the working environment beforehand. 

As stated earlier, we make use of the RatSLAM loop closure detection and map correction back-end in this work \cite{ratslam, ratslamcode}. 
RatSLAM has been proposed as a bio-inspired system following the navigational processes of the rat's hippocampus \cite{ratslam}. Even though the original RatSLAM uses \textit{raw RGB} images for template matching
, further evolutions of RatSLAM, such as LatentSLAM, replace the raw RGB input by the stream of associated latent codes obtained through learned feature extraction \cite{latentslam}. In this work, we feed both our proposed \textit{radar-gyroscope} odometry and the latent codes inferred by our proposed continual learning SNN-STDP fusion system to the RatSLAM back-end.

Since RGB-based SLAMs constitute today's state of the art, we will benchmark our DVS-Radar SLAM 
against both LatentSLAM and RatSLAM, and against ORB features \cite{orbfeature} extensively used in state-of-the-art SLAM systems \cite{orbslam2, orbatlas}. 

\section{SNN Background Theory}
\label{preli}
Unlike frame-based DNNs, SNNs make use of \textit{event-driven} spiking neurons as activation function, often modelled using the Leaky Integrate and Fire (LIF) neurons \cite{lifneuron}:
\begin{equation}
 \begin{cases}
    \frac{dV}{dt} = \frac{1}{\tau_m} (J_{in} - V) \textbf{ with } J_{in} = \bar{w}_{syn}^T \bar{s}(t)
    \\
    \sigma = 1, V \xleftarrow{} 0 \hspace{3pt} \text{\textbf{if}} \hspace{3pt} V \geq \mu, \hspace{3pt} \text{\textbf{else}} \hspace{3pt} \sigma = 0
  \end{cases}
  \label{liff}
\end{equation}
where $\sigma$ is the spiking output, $V$ the membrane potential, $\tau_m$ the membrane time constant, $\mu$ the neuron threshold and $J_{in} = \bar{w}_{syn}^T \bar{s}(t)$ the input to the neuron, resulting from the inner product between the neuron weights $\bar{w}_{syn}$ and the \textit{spiking} input vector $\bar{s}(t)$. The LIF continuously integrates its input $J_{in}$ in $V$ following (\ref{liff}). When $V$ crosses the firing threshold $\mu$, the membrane potential is reset back to zero and an output spike $\sigma = 1$ is emitted. 
SNNs are therefore a natural choice for processing event-driven sensors such as DVS, by \textit{flattening} along space the spiking image in Fig. \ref{dvs_concept} into a vector of spike trains $\bar{s}(t)$. 

\begin{figure}[htbp]
\centering
    \includegraphics[scale = 0.41]{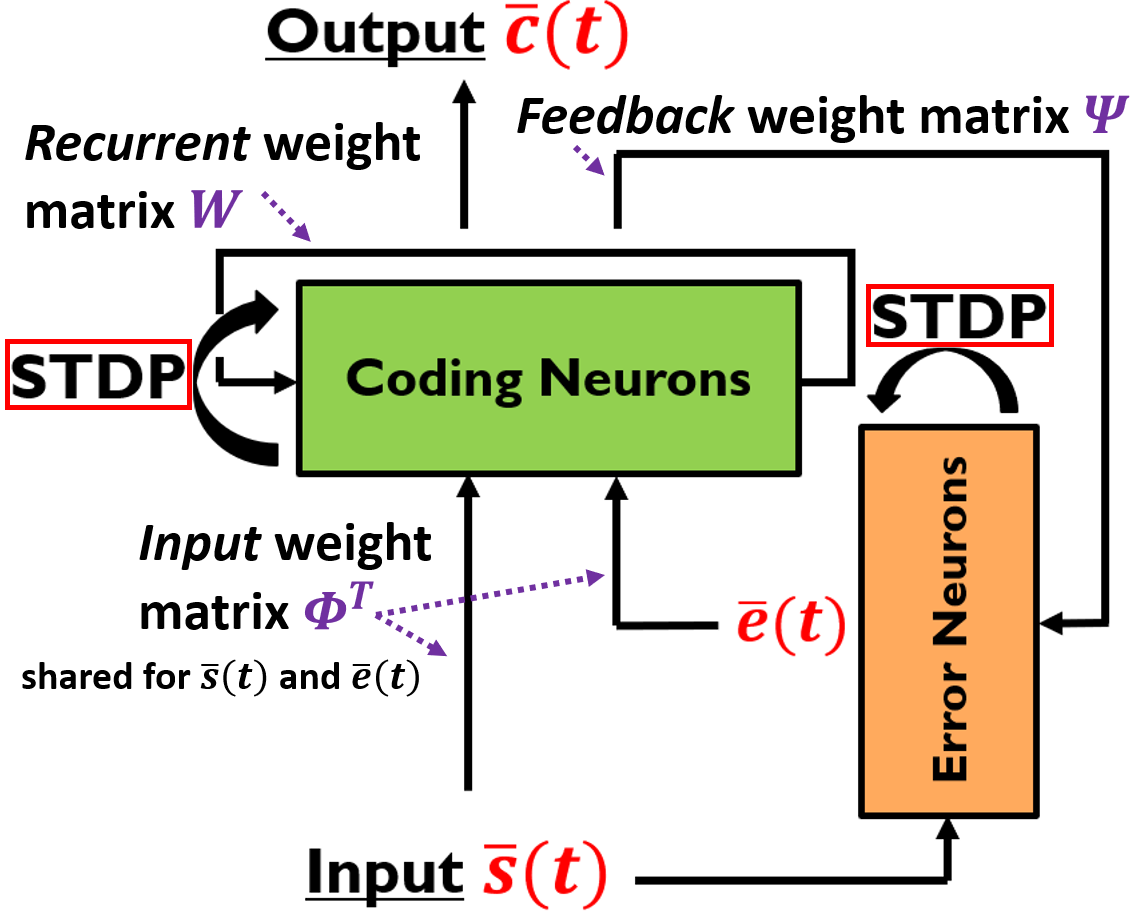}
    \caption{\textit{\textbf{SNN-STDP architecture}. The network is composed of two distinct ensembles of LIF neurons ($M$ \textit{coding} neurons and $N$ \textit{error} neurons). The input spikes $\Bar{s}(t)$ are fed to the coding neurons via the fully-connected weights $\Phi^T$. The coding neurons recurrently infer the SNN output spikes $\bar{c}(t) \xleftarrow{} \text{LIF}\{W \Bar{c}(t) + \Phi^T \Bar{s}(t)\}$ by integrating $\bar{s}(t)$ with the past outputs $\Bar{c}(t)$ through recurrent weights $W$. The \textit{error} neurons receive $\Bar{c}(t)$ via weights $\Psi$ and infer the error spikes $\Bar{e}(t) \xleftarrow{} \text{LIF}\{\Psi \Bar{c}(t) - \Bar{s}(t)\}$ used during STDP learning. STDP (\ref{stdppp})-(\ref{stdppp2}) is locally applied to weights $\Phi, W, \Psi$ following \cite{alistdp}. }}
    \label{snnstdp_orig}
\end{figure}

In order to detect loop closures and perform map correction, learning-based SLAMs seek to infer lower-dimensional latent codes $\bar{c}$ from the sensory observation \cite{latentslam}. 
In this work, we use the SNN in Fig. \ref{snnstdp_orig}, proposed in \cite{alistdp}, as our baseline architecture to continuously infer \textit{spiking} latent codes $\bar{c}(t)$ from the input $\bar{s}(t)$, while jointly \textit{learning} features $\Phi$ from the data in an \textit{unsupervised} way. This can be formulated as:
\begin{equation}
    \Bar{c}, \Phi = \arg \min_{\Bar{c}, \Phi} ||\Phi \Bar{c} - \Bar{s}||_{2}^{2} + \lambda_1 ||\Bar{c}||_1 + \frac{\lambda_2}{2} ||\Phi||_F^2
    \label{lassodefdivc}
\end{equation}
with $\bar{s}$ and $\bar{c}$ the $N$-dimensional input and $M$-dimensional output vectors, with their elements representing the \textit{average spike rates} of the corresponding input and output spike train vectors $\bar{s}(t)$ and $\bar{c}(t)$. $\Phi$ is the transposed SNN input weight matrix, $\lambda_1$ is a sparsity-controlling parameter (defined by the LIF threshold $\mu$ \cite{alistdp}) and $\lambda_2$ is a weight decay parameter. The term $||\Phi \Bar{c} - \Bar{s}||_{2}^{2}$ in (\ref{lassodefdivc}) minimizes the re-projection error between the output latent code $\bar{c}$ and the input $\bar{s}$ through the weights $\Phi$ \textit{without supervision} (similar to an auto-encoder). 

It can be shown \cite{alistdp} that the SNN in Fig. \ref{snnstdp_orig} solves (\ref{lassodefdivc}) by \textit{jointly} inferring $\bar{c}(t)$ while \textit{continuously} adapting its weights $\Phi, W, \Psi$ via STDP learning (\ref{stdppp})-(\ref{stdppp2}) \cite{alistdp}. In (\ref{stdppp}) and (\ref{stdppp2}), $\eta_d$ is the learning rate, $A_p, A_n$ the long-term potentiation (LTP) and depression (LTD) weights, $\tau_p, \tau_n$ the potentiation and depression decay constants, $w_{syn, ij}$ the $j^{th}$ weight of neuron $i$ and $\tau_{ij}$ the time difference between post- and pre-synaptic spike times across the $j^{th}$ synapse of neuron $i$. 
\begin{equation}
w_{syn, ij} \xleftarrow{} w_{syn, ij} + \eta_d \kappa (\tau_{ij})
\label{stdppp}
\end{equation}
\begin{equation}
    \kappa (\tau_{ij})  = \begin{cases}
  A_{p} e^{-\tau_{ij}/ \tau_p}, & \text{if } \tau_{ij} \geq 0
\\
    -  A_{n} e^{\tau_{ij}/ \tau_n}, & \text{if } \tau_{ij} < 0
\end{cases}
\label{stdppp2}
\end{equation}
Eq. (\ref{lassodefdivc}) is never solved explicitly, but rather solved \textit{implicitly} by the SNN-STDP of Fig. \ref{snnstdp_orig} as it iterates. Each weight learns in a local manner, avoiding weight transport problems \cite{weighttransport}.

In Section \ref{archbig}, we use the SNN in Fig. \ref{snnstdp_orig} to build a DVS-Radar fusion architecture that outputs latent codes while continuously learning on the fly \textit{without} any pre-training.





\label{preli}
\label{snninthis}

\section{Proposed method}
\label{props}
\subsection{DVS-Radar fusion architecture using SNN-STDP}
\label{archbig}

\begin{figure}[htbp]
\centering
    \includegraphics[scale = 0.435]{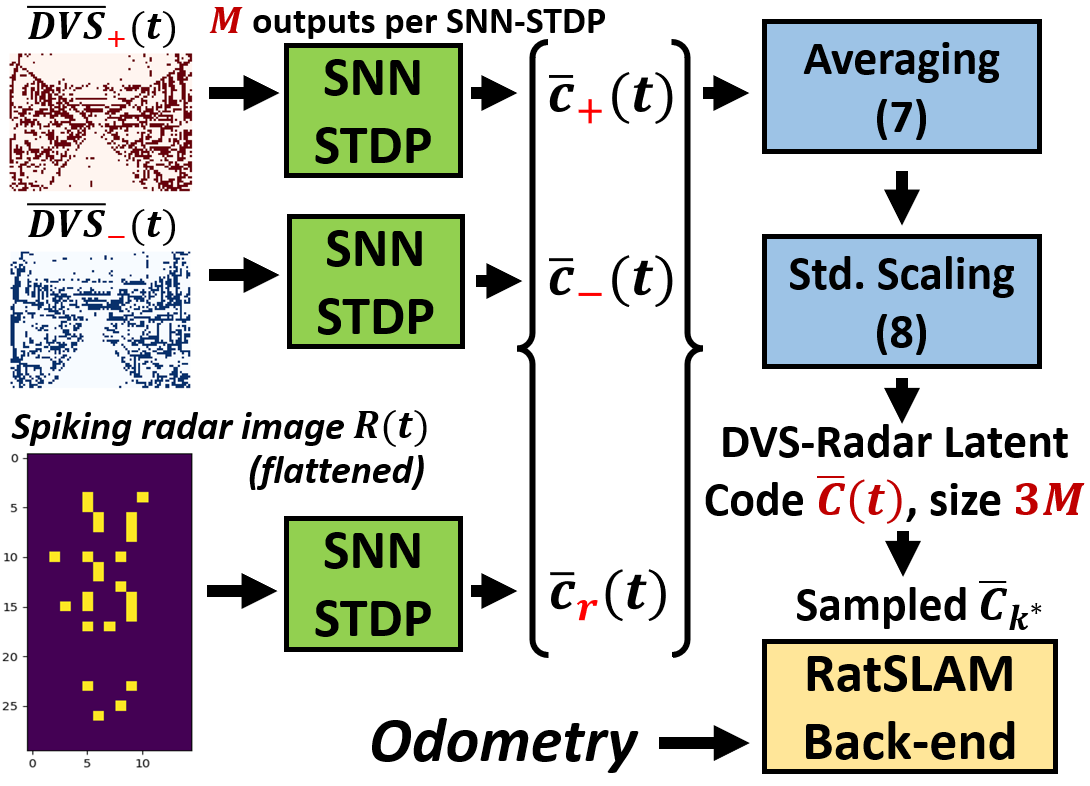} 
    \caption{\textit{\textbf{DVS-Radar SLAM} using SNN-STDP ensembles. The spiking DVS image is flattened as a vector and each event polarity ($+$ or $-$) is fed to its corresponding SNN-STDP network. Radar detections are transformed into a spiking image using (\ref{quantrad},\ref{assignrad}) and flattened before being fed as a spiking vector to the corresponding SNN-STDP. The outputs $\bar{c}_{+,-,r}$ of each SNN are concatenated, then averaged on a time window (\ref{averaged}) and scaled (\ref{stdscaling}) to get $\bar{\mathcal{C}}(t)$. For each time-step $k^*$, the latent codes $\bar{\mathcal{C}}_{k^*}$ and the odometry are fed to a RatSLAM back-end \cite{ratslam} for loop closure detection and map correction. }}
    \label{fusionsnn}
\end{figure}

In order to jointly infer latent codes and continually learn a set of SNN weights capturing the DVS and radar data features, we use the unsupervised  SNN-STDP ensemble of Fig. \ref{snnstdp_orig} (see Section \ref{snninthis}) in a \textit{fusion} setting, by assigning: \textit{i)} a first SNN-STDP ensemble to the \textit{positive} polarity of the DVS; \textit{ii)} a second SNN-STDP ensemble to the \textit{negative} polarity of the DVS, and \textit{iii)} a third SNN-STDP ensemble to the radar detection output (see Fig. \ref{fusionsnn}).

In our experiments, we degrade the time resolution of the DVS camera to spiking bins of $\delta T_{\text{DVS}} = 5$ms and the DVS resolution to $65 \times 86$, as a good balance between computational time and performance. Each polarity channel of the DVS data is then directly connected to its respective SNN-STDP ensemble by flattening the image plane as a binary spiking vector ($\Bar{\text{DVS}}_{+,-}(t)$ in Fig. \ref{fusionsnn}).

For the radar data, our drone setup of Fig. \ref{droneplacehold} 
provides radar detections of the form:
\begin{equation}
    \mathcal{D}_k = \{(x_{d,i}, y_{d,i}, v_{d,i})  \hspace{5pt} \forall i = 1,...,N_t\}
    \label{radardete}
\end{equation}
where $k$ is the radar frame time index, $x_d, y_d$ are the Cartesian coordinates of the detection (in meters), $v_d$ is the radial (or Doppler) velocity of the detection \cite{mydetector}, and $N_t$ is the number of detected targets in frame $k$ (with a frame rate of $20$ FPS). Therefore, radar detections can be seen as binary events (or spikes) with their address given by the associated coordinate $(x_d, y_d)$. 

We transform the detection set (\ref{radardete}) into a \textit{spiking image} $R$ of size $W_r \times H_r$ by \textit{a)} quantizing the coordinates to lie between $[0, W_r-1]$ for $x_d$ and $[0, H_r-1]$ for $y_d$ via (\ref{quantrad}), and \textit{b)} assigning a value of $1$ to the radar image pixels associated with a detection coordinate and $0$ everywhere else, via (\ref{assignrad}).
\begin{equation}
    Q^x_{d,i} = \lfloor (W_r - 1) \frac{x_{d,i}}{\text{max}_x} \rfloor \hspace{5pt} Q^y_{d,i} = \lfloor (H_r - 1) \frac{y_{d,i}}{\text{max}_y} \rfloor
    \label{quantrad}
\end{equation}
\begin{equation}
    R[Q^x_{d,i}, Q^y_{d,i}] = 1 \hspace{5pt} \forall i = 1,...,N_t \hspace{5pt} \textbf{else} \hspace{5pt} R = 0
    \label{assignrad}
\end{equation}
where $\text{max}_x = 10$m and $\text{max}_y = 10$m are the maximum coordinate extents that the radar can detect. We set $W_r=15, H_r = 30$ as a good balance between radar image resolution and computational time during SNN processing. An example of a \textit{spiking radar image} obtained using (\ref{quantrad})-(\ref{assignrad}) is shown in Fig. \ref{fusionsnn}. 
Next, we feed the \textit{spiking radar image} $R$ to its SNN-STDP ensemble by flattening $R$ as a spiking vector with an oversampling in time of $\times 10$ in order to meet the DVS time resolution of $\delta T_{\text{DVS}}=5$ms, i.e., each spiking radar image is fed $10$ times to the SNN-STDP input.

As the drone navigates, DVS and radar data are fed to the SNN ensembles which continuously learn features via STDP and infer $M$-dimensional \textit{spiking} outputs $\bar{c}_{+}(t), \bar{c}_{-}(t), \bar{c}_r(t)$ corresponding to the \textit{positive} DVS channel, the \textit{negative} DVS channel and the radar data. The outputs are then concatenated and averaged on a time window of length $\Delta T_{\text{avg}} = 0.1$s to obtain the fused latent code $\bar{\mathcal{C}} \in \mathbb{R}^{3M}$ (see Fig. \ref{fusionsnn}):
\begin{equation}
  \bar{\mathcal{C}}(t^*) = \frac{1}{\Delta T_{\text{avg}}}\int_{t^*-\Delta T_{\text{avg}}}^{t^*} \begin{bmatrix}
           \bar{c}_{+}(t) \\
           \bar{c}_{-}(t) \\
            \bar{c}_r(t)
         \end{bmatrix}
         dt
  \label{averaged}
\end{equation}

Finally, we apply \textit{standard scaling} to compensate for the covariate shifts during drone navigation \cite{batchnorm}, as well as for the scale differences between the elements of $\bar{\mathcal{C}}$, greatly helping template matching between the latent codes:
\begin{equation}
    \mathcal{C}_l \xleftarrow{} \frac{\mathcal{C}_l - m_l}{std_l}
    \label{stdscaling}
\end{equation}
where the mean $m_l$ and standard deviation $std_l$ of element $\mathcal{C}_l$ are estimated \textit{on-line} using the algorithm provided in \cite{stdscaling}. 

Fig. \ref{fusionsnn} shows that the \textit{averaged} and \textit{scaled} latent codes $\bar{\mathcal{C}}(t)$, together with the raw \textit{radar-gyroscope} odometry (see Section \ref{odom}) are fed to the RatSLAM back-end \cite{ratslam} (at a rate $f_s=10$ FPS). 
The RatSLAM back-end detects loop closures by measuring the similarity between the currently sampled latent code $\bar{\mathcal{C}}_{k^*}$ and the previous latent codes $\bar{\mathcal{C}}_{k}, \forall k < k^*$ associated to the \textit{past} drone poses in order to correct all drone poses $\{X_k, Y_k, \Psi_k\}$ (with location $X_k, Y_k$ and yaw angle $\Psi_k$), maintaining localisation and mapping.

\subsection{Radar-Gyroscope Odometry}
\label{odom}
Since accelerometer data can be too noisy due to the erratic drone vibrations, we use the radar sensor to retrieve the drone heading velocity $v_{h}$, while the on-board gyroscope is used to measure the yaw rotation velocity $\dot{\Psi}$. The odometry is obtained by integrating $v_h$ and $\dot{\Psi}$ following the standard equations of movements:
\begin{equation}
 \begin{cases}
    \Psi_k \xleftarrow{} \Psi_{k-1} + \dot{\Psi} \delta t_{g}
    \\
    X_k \xleftarrow{} X_{k-1} + v_h \cos(\Psi_k) \delta t_{r}
    \\
    Y_k \xleftarrow{} Y_{k-1} + v_h \sin(\Psi_k) \delta t_{r} 
  \end{cases}
  \label{odo_integ}
\end{equation}

where $k$ denotes the time step or pose index, $\delta t_g = 1$ms is the gyroscope sampling period, $\delta t_r = 40$ms is the radar frame period, and $X_k,Y_k$ denote the drone coordinates. 

We measure the drone heading velocity $v_h$ from the radar detections in each frame by first projecting the Doppler velocity \cite{mydetector} of each detection back to the heading direction. Then, we take the median value of all the obtained velocities following Algorithm \ref{radarodomalg}. We use the median value instead of the mean in order to provide robustness to outliers. In addition, we restrict the detections to lie in a $\pm 60^\circ$ field of view around the drone heading axis in order to further reject unreliable measurements that are caused by the attenuation of the radar antenna gain profile for high azimuth angles. 
\begin{algorithm}
 \caption{Radar odometry for heading velocity}
 \label{radarodomalg}
 \begin{algorithmic}[1]
 \renewcommand{\algorithmicrequire}{\textbf{Input:}}
 \renewcommand{\algorithmicensure}{\textbf{Output:}}
 \REQUIRE Radar detections: $\{x_{d,i}, y_{d,i}, v_{d,i}  \hspace{5pt} \forall i = 1,...,N_t\}$ for the current frame, Valid angle range $\theta_v=60^\circ$.
 \ENSURE  Heading velocity: $v_h$ 
 \STATE $V = \{\}$: empty array
  \FOR {$i \in \{1,...,N_d \}$}
  \STATE // Loop over all detections in the current radar frame
  \STATE $\theta_i = |\arctan(\frac{x_{d,i}}{y_{d,i}})|$
  \IF {$\theta_i < \theta_v$}
  \STATE // If detection angle is within the valid range
  \STATE Append $\frac{v_{d,i}}{\cos{\theta_i}}$ to $V$ //projecting radial velocity to
  \ENDIF \hspace{60pt} y-axis (i.e., drone heading axis).
  \ENDFOR
  \RETURN $v_h = \text{median}(V)$ 
 \end{algorithmic} 
 \end{algorithm}
 

\subsection{Radar-based obstacle aggregation}
\label{obstacleaggreg}
The radar sensor provides a direct and robust mean for the reconstruction of walls and other landmarks observed by the drone during its flight. For each pose $k$, a set of radar detections $\mathcal{D}_k = \{x_{d,i}, y_{d,i}  \hspace{5pt} \forall i\}$ is available and can be aggregated with the previous detections and the previous drone poses in order to reconstruct the walls and obstacles. For each pose $k$, the radar detections are rotated by the drone yaw angle $\Psi_k$ and translated by the drone position $[X_k, Y_k]$:
\begin{equation}
    \begin{bmatrix}
\tilde{x}_{d,i} \\
\tilde{y}_{d,i} 
\end{bmatrix} \xleftarrow{}  
    \begin{bmatrix}
\cos(\Psi_k) & -\sin(\Psi_k) \\
\sin(\Psi_k) & \cos(\Psi_k)
\end{bmatrix} 
 \begin{bmatrix}
x_{d,i} \\
y_{d,i} 
\end{bmatrix} +
 \begin{bmatrix}
X_k \\
Y_k 
\end{bmatrix} \hspace{5pt}
\forall i
\label{wallrottrans}
\end{equation}

Therefore, each time a loop closure is detected (see Section \ref{archbig}), our pipeline refines not only the past poses $\{X_k, Y_k, \Psi_k\}$, but also the obstacle coordinates $[\tilde{x}_{d,i}, \tilde{y}_{d,i}]$ by applying (\ref{wallrottrans}) again with the corrected $\Psi_k$ and $[X_k, Y_k]$. Our obstacle aggregation is shown in both Fig. \ref{droneplacehold} b) and Fig. \ref{seq3}, where walls and obstacles can clearly be seen.

\section{Experimental Results}

\begin{figure}[htbp]
\centering
    \includegraphics[scale = 0.37]{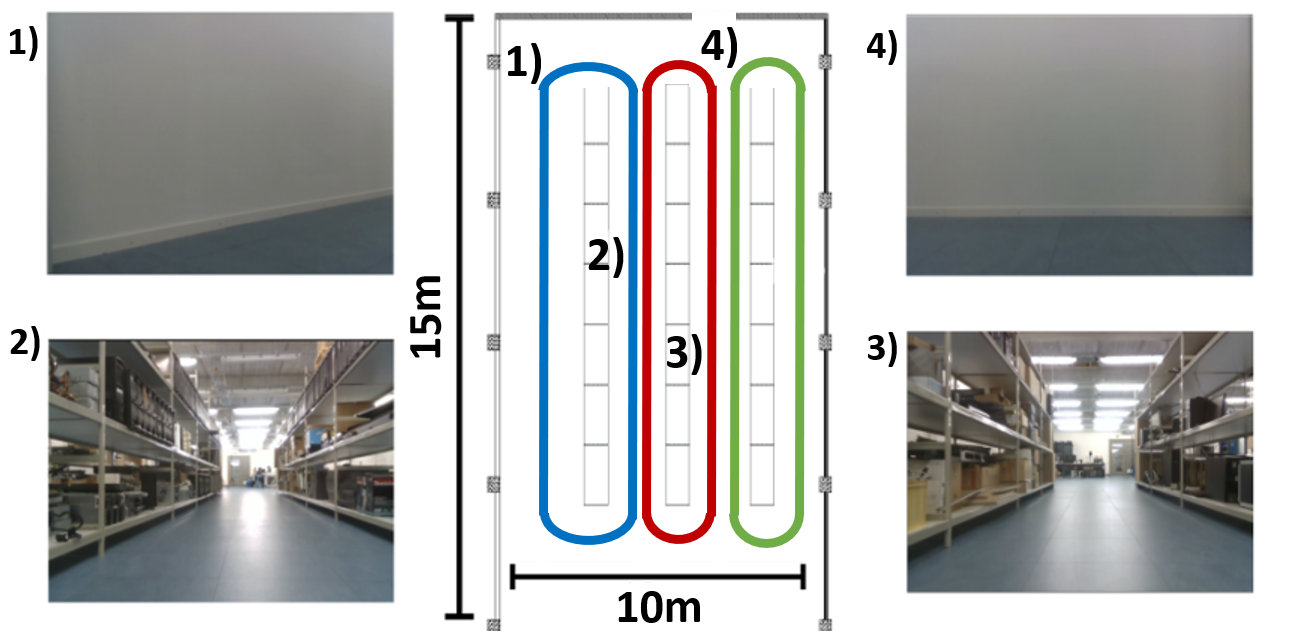}
    \caption{\textit{\textbf{Flight data used in this work.} The drone location is hard to disambiguate visually (e.g., view 1 vs. 4, 2 vs. 3). Data were acquired during three different flight sequences: seq.1: the red path; seq.2: a combination of all paths, and seq.3: the blue and red path with strong lighting variations.}}
    \label{aliasing}
\end{figure}

We will now compare the \textit{mean absolute error} (MAE) of our method against state-of-the-art RGB-based solutions on three different flight sequences (Seq.1-2-3 in Fig. \ref{aliasing}). In addition, we test our DVS-Radar system under lighting variations (Seq.3 in Fig. \ref{aliasing}), by randomly switching the lights on and off. Table \ref{paramsdvstrain} reports the SNN-STDP parameter values used in our experiments, tuned manually for optimising the SLAM performance, starting from the values in \cite{alistdp}. 
\begin{table}[h]
\centering
\begin{tabular}{|c|c|c|c|c|c|c|c|}
\hline
$\boldsymbol{\eta_d}$ & $\boldsymbol{\eta_c}$ & $\boldsymbol{M}$ & $\boldsymbol{N_{+,-}}$ & $\boldsymbol{N_r}$ & $\boldsymbol{\mu_{+,-}}$ & $\boldsymbol{\mu_{r}}$  \\
\hline
$3\times 10^{-4}$ & $1$ & $64$ & $65 \times 86$ & $15 \times 30$ & $0.3$ & $0.03$ \\
\hline
\end{tabular}

\begin{tabular}{|c|c|c|c|c|c|c|}
$\boldsymbol{\tau_{m}^{+,-}}$ & $\boldsymbol{\tau_{m}^{r}}$ & $\boldsymbol{A_p}$ & $\boldsymbol{A_n}$ & $\boldsymbol{\tau_p}$ & $\boldsymbol{\tau_n}$  \\
\hline
$0.017$ s & $0.17$ s & $1$ & $0.8$ & $0.0208$ s & $0.008$ s  \\
\hline
\end{tabular}
\caption{\textit{\textbf{SNN-STDP parameters.} $\eta_d$ is the STDP learning rate. $\eta_c$ is the SNN coding rate \cite{alistdp}. The number of coding neurons $M$ is the same for all SNN ensembles in Fig. \ref{fusionsnn}. The number of error neuron $N_{+,-,r}$ is equal to the dimensions of the input image to each SNN, where $+,-,r$ respectively denote the SNN ensemble for the positive DVS, the negative DVS and the radar data in Fig. \ref{fusionsnn}.  $\mu_{+,-,r}$ and $\tau_m^{+,-,r}$ are the neuron thresholds and membrane constants in (\ref{liff}) for the respective SNNs. $A_{p,n}$, $\tau_{p,n}$ are the STDP parameters in (\ref{stdppp2}). 
}}
\label{paramsdvstrain}
\end{table}

Data is acquired in a challenging warehouse environment composed of storage aisles, with high visual ambiguities between the scenes. This makes the disambiguation of the drone location hard due to the redundancy between the different views \cite{latentslam, alias} (see Fig. \ref{aliasing}).
\label{expres}
Our warehouse \cite{indooruwb} is also equipped with \textit{Ultra Wide Band} (UWB) beacons for ground truth positioning, noted $(x_{k}^{\text{gt}}, y_{k}^{\text{gt}})$ with $k$ the time index, \textit{only} used for error assessment. When computing the error, since the ground truth coordinates and the SLAM coordinates are not in the same coordinate system \cite{ratslam}, the ground truth coordinates are \textit{translated} and \textit{rotated} in order to match the SLAM coordinate system. We identify the rotation angle and translation vector by minimizing for the drone \textit{localisation} MAE (identified via grid search). This \textit{localisation} MAE is obtained as the average error between the ground truth $(x_{k}^{\text{gt}}, y_{k}^{\text{gt}})$ and the SLAM localisation $(x_{k}, y_{k})$ on the complete flight sequence (with length $T_{end}$):
\begin{equation}
    \text{MAE}_{\text{L}} = \frac{1}{T_{end}} \sum_{k=1}^{T_{end}} (|x_{k} - x_k^{\text{gt}}| + |y_{k} - y_k^{\text{gt}}|)
    \label{locerr}
\end{equation}

In addition, the \textit{mapping} MAE is obtained \textit{a posteriori}, as the deviation between \textit{all} ground truth points and the map obtained at the end of the SLAM process:
\begin{equation}
        \text{MAE}_{\text{M}} = \frac{1}{T_{end}} \sum_{k=1}^{T_{end}} \min_j \{ |x_{k} - x_j^{\text{gt}}| + |y_{k} - y_j^{\text{gt}}| \}
        \label{maperr}
\end{equation}

\subsection{Normal lighting conditions}
Table \ref{performanceloc} reports the $\text{MAE}_{\text{L}}$ and $\text{MAE}_{\text{M}}$ of our approach against state-of-the-art solutions that make use of standard RGB camera data: \textit{i)} the original RatSLAM \cite{ratslam}; 
\textit{ii)} the \textit{LatentSLAM}, trained specifically for our environment following \cite{latentslam}, on a set of 3998 frames acquired \textit{independently} from Seq-1-2-3; and \textit{iii)} ORB features for template matching \cite{orbfeature} using \textit{Lowe's Ratio Test} \cite{lowe}. Table \ref{performanceloc} also reports ablation results when using only one modality alone.

For LatentSLAM and our method, the similarity between two frames $i,j$ is computed as $s_{ij} = \frac{\bar{ \mathcal{C}}_i^T \bar{ \mathcal{C}}_j}{||\bar{ \mathcal{C}}_i ||_2 || \bar{ \mathcal{C}}_j||_2}$ where $\bar{ \mathcal{C}}_{i,j}$ is the latent code (\ref{averaged}, \ref{stdscaling}) obtained by each method. 

For ORB feature matching, $s_{ij}$ is computed as $s_{ij} = 1- N_{m} / \max_{m}$ with $N_m$ the number of matches and $\max_{m}$ the maximum number of matches between the two frames $i,j$. 

The RatSLAM back-end integrates the odometry of Section \ref{odom} and detects \textit{loop closures} by sampling the latent codes $\bar{ \mathcal{C}}_{i}$ at $100$ms intervals and by testing the similarities $s_{ij}, \forall j$ against a threshold $\theta$ tuned for minimum MAE ($\theta=0.65$ for Radar-only, $\theta=0.6$ for DVS-only and DVS-Radar). 

\begin{table}[htbp]
\begin{tabularx}{0.48\textwidth}{@{}l*{3}{c}c@{}}
\toprule
Architecture  &  Seq-1 (L) &  Seq-1 (M) &  Seq-2 (L) &  Seq-2 (M)  \\ 
\midrule
RatSLAM \cite{ratslam}   &  0.72         &   0.22   &  2.49        &   0.57       \\ 
LatentSLAM \cite{latentslam}  &  0.54       &   0.21     &  1.68        &   0.39      \\
ORB features \cite{orbfeature}  & \underline{0.46}         &  \underline{0.17}      &  0.95        &   \underline{0.39}        \\
\textbf{Ours} (Radar)   & 0.68      &  0.28   &  0.98        &  0.64 \\ 
\textbf{Ours} (DVS)   &  1.65       &  0.19  & 0.84     &   0.51  \\
\textbf{Ours} (DVS-Radar)   & 0.51      &  \underline{0.17}   &  \underline{0.81}     &   0.45  \\
\bottomrule
\end{tabularx}
\caption{\textit{\textbf{$\text{MAE}_{\text{L}}$ (L) and $\text{MAE}_{\text{M}}$ (M).} The lower the better. }
}
\label{performanceloc}
\end{table}

\begin{figure}[htbp]
\centering
    \includegraphics[scale = 0.41]{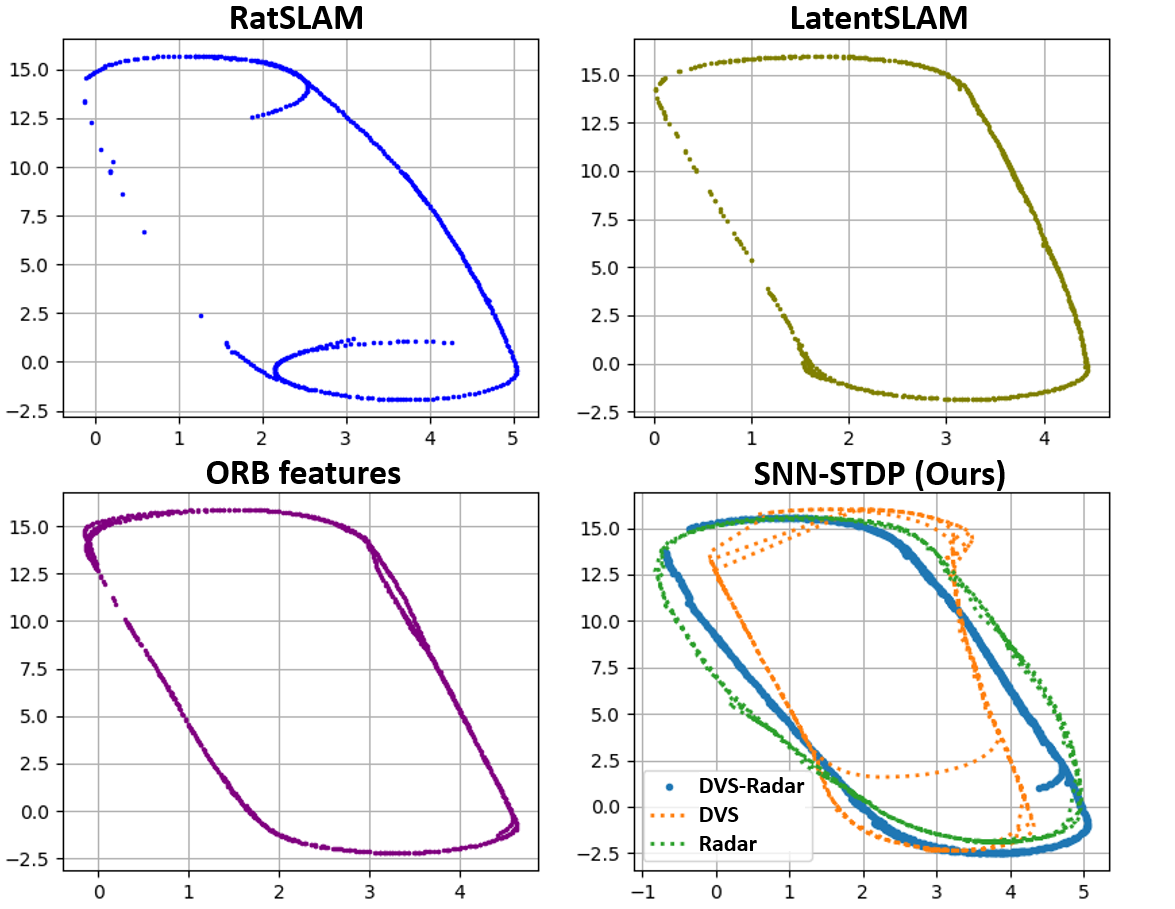}
    \caption{\textit{\textbf{Seq-1 (dimensions in meter)}, red path in Fig. \ref{aliasing}  (the obstacle modelling of Section \ref{obstacleaggreg} is not plotted for clarity). }}
    \label{seq1}
\end{figure}

\begin{figure}[htbp]
\centering
\vspace{5pt}
    \includegraphics[scale = 0.41]{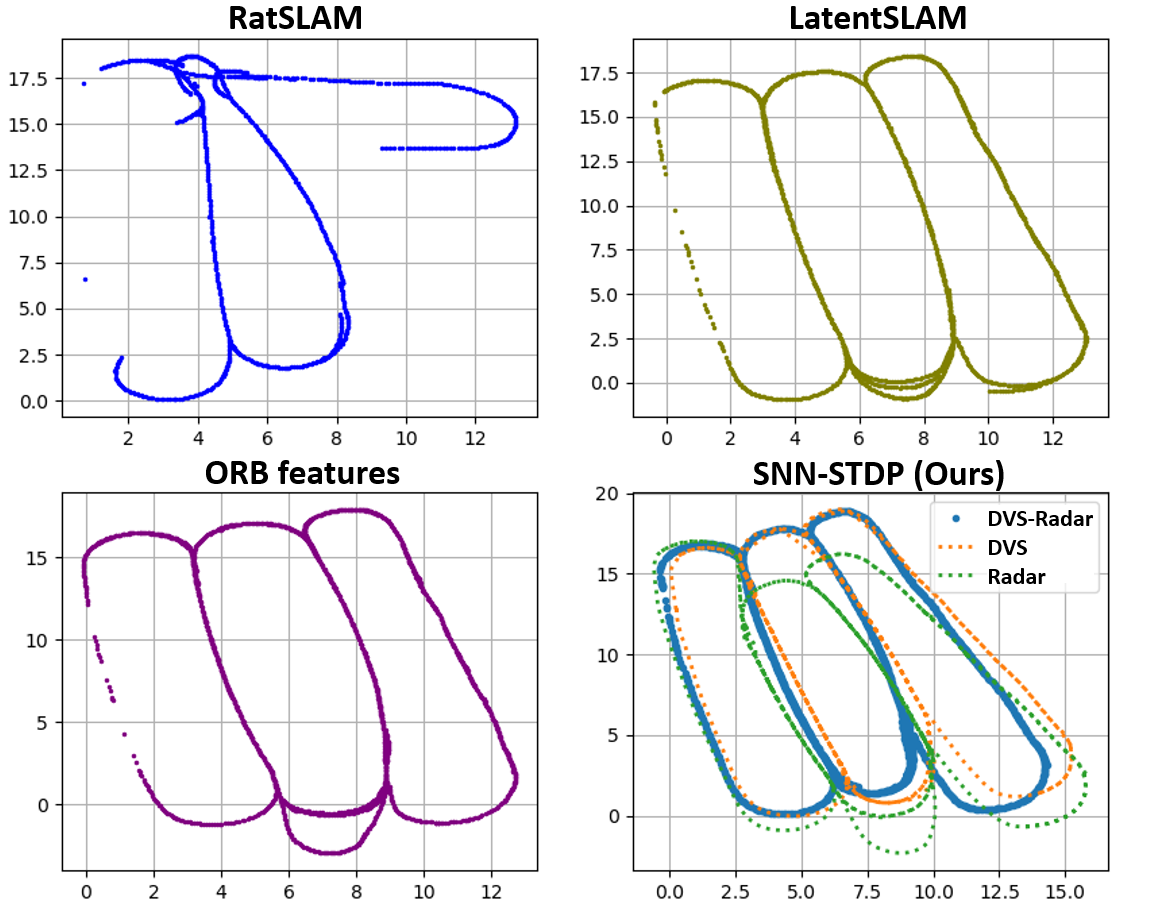}
    \caption{\textit{\textbf{Seq-2}, all paths combined in Fig. \ref{aliasing} (the obstacle modelling of Section \ref{obstacleaggreg} is not plotted for clarity).}}
    \label{seq2}
\end{figure}

In addition to Table \ref{performanceloc}, Fig. \ref{seq1} and \ref{seq2} visually show the SLAM results for both Seq-1 and Seq-2. It can be remarked on both Fig. \ref{seq1}, \ref{seq2} and in Table \ref{performanceloc} that the DVS-Radar fusion setup outperforms the Radar-only and the DVS-only cases, by reaching lower $\text{MAE}_{\text{L}}$ and $\text{MAE}_{\text{M}}$ values. This clearly shows the advantage of DVS-Radar fusion.

Compared to the previously-proposed RGB-based solutions, our proposed system outperforms both the original RatSLAM and the LatentSLAM, which makes use of a 11-layer DNN trained \textit{offline}. In addition, our system either outperforms or reaches close $\text{MAE}_{\text{L}}$ and $\text{MAE}_{\text{M}}$ performances compared to ORB features used with RGB.

This is \textit{remarkable} given \textit{a)} the small size of our SNN-STDP networks (equivalent to a 1-hidden-layer network in terms of complexity versus 11 layers in LatentSLAM), \textit{b)} the fact that our network is not pre-trained offline as in LatentSLAM, but is rather initialized randomly and learns features from the input data \textit{on the fly} via \textit{unsupervised} continual STDP adaptation, and \textit{c)} the fact that it does not use RGB as in most SLAMs and solely relies on \textit{lower-fidelity} DVS and Radar data. In contrast to the RGB-based solutions of Table \ref{performanceloc}, our DVS-Radar setup enables us to work under strong lighting variations, where RGB solutions fail.



\subsection{Under low light conditions}
\label{degradedlight}

\begin{figure}[htbp]
\centering
    \includegraphics[scale = 0.42]{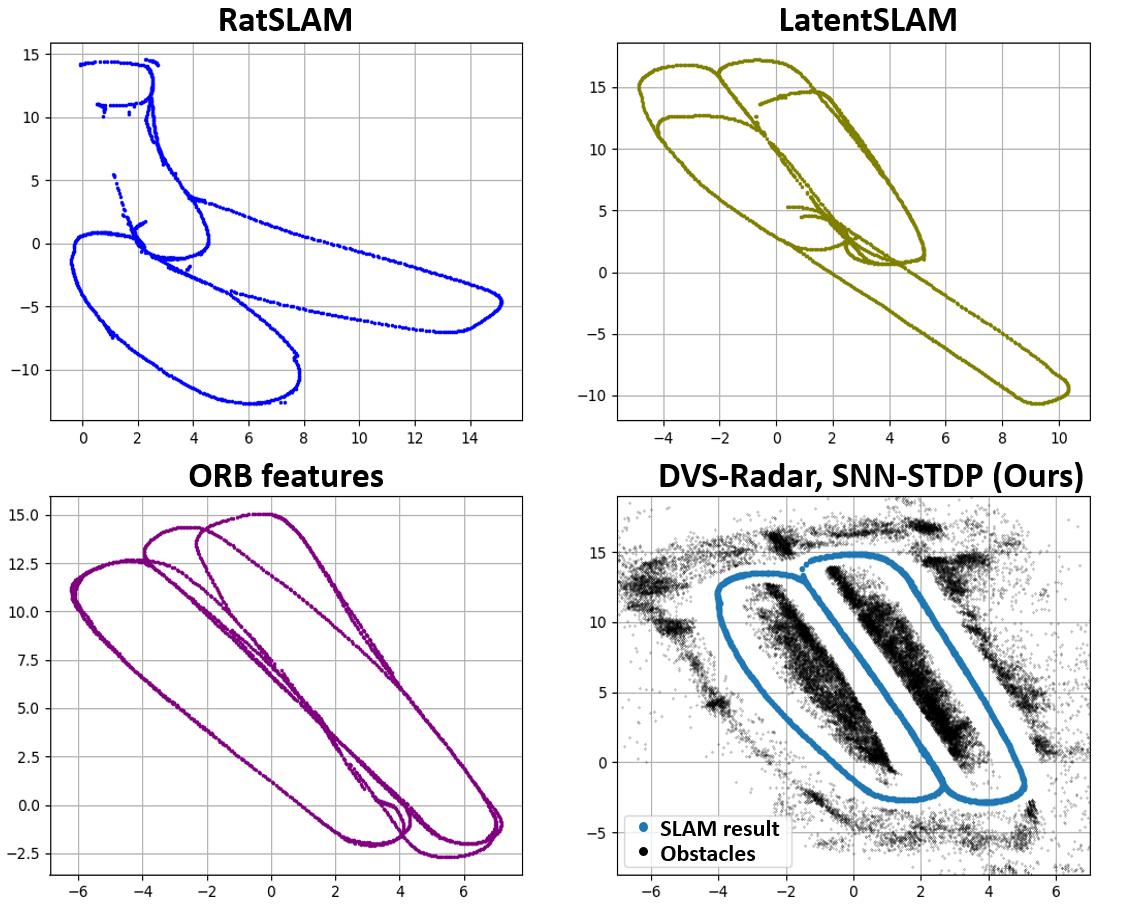}
    \caption{\textit{\textbf{Seq-3}, blue and red paths in Fig. \ref{aliasing} with strong lighting variations. Our proposed DVS-Radar system is robust against lighting and clearly outperforms the other methods. Our obstacle aggregation of Section \ref{obstacleaggreg} is also shown, where walls and shelves are modelled by the black dots. }}
    \label{seq3}
\end{figure}

Fig. \ref{seq3} shows the SLAM results obtained for Seq-3 of Fig. \ref{aliasing}, where lighting variations were conducted by randomly turning on and off a subset of the neon lighting in the warehouse during the drone flight. Fig. \ref{seq3} clearly demonstrates the \textit{robustness} of our DVS-Radar approach in low light, compared to the other methods. 

It must be remarked that systems fusing RGB with radar have been proposed for having robustness towards environmental conditions \cite{latentslam, rgbradarfusion}. However, the RGB sensor still remains unreliable in low lighting, while in our architecture, both the DVS and radar stay functional under low light. 
A video showcasing our system on Seq-3 is provided at \url{https://youtu.be/a7gvZWNHGoI}.


\section{Conclusions}
\label{concs}
This paper has presented what is, to the best of our knowledge, a first-of-its-kind SLAM system fusing DVS and radar data with SNNs. Unlike most learning-based SLAM systems, where a DNN is trained offline using a dataset captured beforehand, our proposed system does not require any pre-training but relies on the continual and \textit{unsupervised} adaptation of the SNN weights via STDP learning as the drone explores the environment. Finally, it has been shown that our DVS-Radar SLAM reports a competitive performance compared to state-of-the-art RGB-based solutions while enabling robust navigation and obstacle modelling under strong lighting variations. 
We believe that this work holds clear promise for the deployment of environmentally robust, continual learning SLAM systems for drones via low-power SNN processors. 

\section*{Acknowledgment}
We thank Prof. J. Suykens of \textit{KU Leuven} and dr. L. Keuninckx and D. de Tinguy of \textit{imec} for useful discussions.


\end{document}